\documentclass[conference]{IEEEtran}
\IEEEoverridecommandlockouts
\usepackage{cite}
\usepackage{amsmath,amssymb,amsfonts}
\usepackage{algorithmic}
\usepackage{graphicx,subcaption} 
\usepackage{textcomp}
\usepackage{bbold}
\usepackage{array}

\newcommand{\real}{\mathbb{R}}
\newcommand{\norm}[1]{\left\lVert#1\right\rVert}
\usepackage{stfloats}
\usepackage{url}
\usepackage{cite}
\usepackage{lineno,hyperref}
\modulolinenumbers[5]
\usepackage[dvipsnames]{xcolor}
\usepackage{mathtools}
\usepackage{enumerate}
\usepackage{booktabs}
\usepackage{enumerate}
\usepackage{booktabs}
\usepackage{color}

\def\BibTeX{{\rm B\kern-.05em{\sc i\kern-.025em b}\kern-.08em
    T\kern-.1667em\lower.7ex\hbox{E}\kern-.125emX}}
\begin{document}

\title{Deep Equilibrium Models Meet Federated Learning}

\author{
\IEEEauthorblockN{Alexandros Gkillas$^{1}$, Dimitris Ampeliotis$^2$, Kostas Berberidis$^1$}
\IEEEauthorblockA{$^1$Dept. of Computer Engineering and Informatics,  University of Patras, Greece\\
$^2$Dept. of Digital Media and Communication, Ionian University, Greece
}\thanks{This  work  was  supported in part by the University of Patras  and the  RPF,  Cyprus, under  the  project  INFRASTRUCTURES/1216/0017  (IRIDA).}
}

\maketitle

\begin{abstract}
In this study the problem of Federated Learning (FL) is explored under a new perspective by utilizing the  Deep Equilibrium (DEQ) models instead of conventional deep learning networks.
We claim that incorporating DEQ models into the federated learning framework naturally addresses several open problems in FL, such as the communication overhead due to the sharing large models and the ability to incorporate heterogeneous edge devices with significantly different computation capabilities.
Additionally, a weighted average fusion rule is proposed at the server-side of the FL framework to account for the different qualities of models from heterogeneous edge devices. 
To the best of our knowledge, this study is the first to establish a connection
between DEQ models and federated learning, contributing to the development of an efficient and effective FL framework. Finally, promising initial experimental results are presented, demonstrating the potential of this approach in addressing challenges of FL.

\end{abstract}


\section{Introduction}

Federated learning (FL) has emerged as a promising approach for privacy-preserving deep learning by distributing both data collection and model training to the edge. In FL, a group of edge devices, e.g., IoT devices, collaboratively optimize deep learning models without sharing any information about their data. Instead of sending their data, the clients train their models locally and periodically send model updates to a central server for aggregation \cite{mcmahan2017communication}. 
In real-world applications FL  encounters two significant challenges, namely, the communication burden between the server and the edge devices and the heterogeneity of the devices in terms of computational and power resources \cite{open_problems, eldar, Asynch_fl_survey}. Recent studies in federated learning have disregarded these two significant limitations \cite{intro_federated}, utilizing large homogeneous deep neural networks. However, in practical IoT settings, the devices are characterized by limited computational and communication resources, which force them to train smaller and computationally lighter neural networks, thus affecting heavily their performance  \cite{Asynch_fl_survey}. 


To increase communication efficiency in federated learning, compression schemes, e.g., sparsification \cite{pmlr-v119-rothchild20a}, quantization \cite{9054168} and client selection \cite{9237168}, have been widely explored. However, these approaches may result in loss of accuracy and introduce bias towards certain devices.
Also, focusing on the hardware heterogeneity of the devices, 
one straightforward approach \cite{MLSYS2019_bd686fd6} is to select only clients with adequate computational resources, while disregarding those with limited hardware, which may still possess valuable information. Alternatively, a model architecture could be employed to fit the minimum capabilities of all clients, but this may constrain the overall representation ability of the global model \cite{Asynch_fl_survey}.
Another direction relies on deploying  different models  across clients adapted to their computational resources. To exchange information over heterogeneous models the knowledge distillation technique is applied 
to enhance the global model with an ensemble of local predictions \cite{MLSYS2019_bd686fd6}.
However, implementing such approaches can be challenging due to the complex aggregation rules required on the server or the need for clients to share a public proxy dataset, which may not be feasible for devices with limited memory \cite{Heter_nips, Heter_4}.

\textbf{Contribution:} Unlike the existing literature, in this study, we examine the FL under a different perspective, focusing on the structure and properties of the model employed by the edge devices. To be more precise, rather than employing a conventional deep learning network as referenced in prior work \cite{eldar}, we utilize the Deep Equilibrium (DEQ) models \cite{NEURIPS2019_01386bd6}. We argue that these models are characterized by unique properties providing solutions to open-problems in FL including the communication burden between the devices and server and   the computational heterogeneity of the local devices.  

Although, the Deep equilibrium models have been explored in numerous centralized settings \cite{pami, chen2021semialgebraic, 10095286, NEURIPS2020_3812f9a5, my}, to the best of our knowledge, our study is the first to investigate a connection between DEQ models and federated learning. More specifically, it is shown here that  expressing the entire  architecture of a deep learning model as an equilibrium (fixed-point) computation of a single layer or unit (e.g., a residual block) results in an efficient infinite-depth neural network  that can offer substantial benefits to federated learning. This compressed representation requires notably less memory, thus  enabling efficient  communication of model updates between the  server and the devices, while achieving
competitive  performance.
Furthermore, the complexity of the DEQ model can adapt dynamically based on the edge devices' computational capabilities by adjusting the number of fixed point iterations required to compute the corresponding equilibrium point. Finally, we propose a novel, weighted average fusion rule that takes into account heterogeneous edge devices that employ different numbers of fixed point iterations, thereby utilizing this information effectively. 
Note that our proposed method is applicable to any federated learning algorithm.

\section{Preliminaries- Deep equilibrium (DEQ) models}\label{deq_pre}

The DEQ model is inspired by the observation that a typical k-layer neural network (NN) (with an input-skip connection) \cite{pami} can be formulated as follows 
\begin{align}
\small
    &z_{i+1} = \sigma(\textbf{W}_i z_i + U_ix +b_i), \quad i=1,\cdot,k-1
\end{align}
where $\textbf{W}_i, \textbf{U}_i$ denote the weight matrices, $b_i$ is  the bias term of the $i-$th layer, $\sigma(\dot)$ corresponds to the activation function, and $x$ is the input. 
Under the weight-tying practice \cite{NEURIPS2019_01386bd6},  $\textbf{W}_i, \textbf{U}_i$ and $b_i$ of each layer can be replaced by the same  $\textbf{W}, \textbf{U}$ and   $b$, thus deriving a weight-tied deep network as follows
\begin{align}
\small
    &z_{i+1} = \sigma(\textbf{W} z_i +\textbf{U}x + b), \quad i=1,..,k-1
    \label{eq:weight_tied_model}
\end{align}
Note that typically a separate weight $\textbf{W}_k$ and bias term $b_k$ is employed to generate the final output, since the output can be a different size than the hidden unit. Focusing on the key iteration in   (\ref{eq:weight_tied_model}), 
the study in \cite{NEURIPS2019_01386bd6} recognised that if we apply the transformation (\ref{eq:weight_tied_model}) for an infinite number of times the output of this transformation should be a fixed point, i.e.,
\begin{equation}
\small
    z^{\star}= \sigma(\textbf{W} z^{\star} +\textbf{U}x + b).
     \label{eq:key_iter_zstar}
\end{equation}
In other words, the DEQ models aim to estimate the fixed point $z^\star$ where any further application of transformation (\ref{eq:weight_tied_model}) would not alter its value. \textit{Note that the above solution corresponds to an \textbf{infinite-depth} network}. In light of this, a DEQ model is defined as a fixed point function $f_\theta(z,x)= \sigma(\textbf{W}z+\textbf{U}x+b)$, which obeys the following relation
\begin{equation}
\small
z^\star = f_\theta(z^\star, x).
\label{eq:fixed_point}
\end{equation}
where $\theta$ denotes the parameters of the model.


\section{Proposed Method}

\subsection{Federated Deep Equilibrium Learning}

To mathematically formulate the considered  federated  deep equilibrium learning problem, we define a set of $\mathcal{N}$ edge devices, where each device $n \in \mathcal{N} = \{1, 2, \ldots , N\}$ contains a local private dataset, denoted as  $\mathcal{D}_n = \{x_{i,n}, y_{i,n} \}_{i=1}^{p_n} $,
where $x_n^i$ is the input, and $ y_n^i$ is the corresponding label.  Given $\mathcal{D}_n$, each device $n$ aims to train a local DEQ model, whose weights are denoted as $\theta_n$. This can be achieved by  minimizing a local objective $g_n(\theta;\mathcal{D}_n)$ that utilizes some loss function, denoted as $\mathcal{L(.)}$. Since, an infinite depth neural network is considered, this minimization process requires the solution of several fixed point problems, needed for the calculation of the output of the DEQ model. In particular, the local objective of device $n$ is 
\begin{equation}
\small
g_n(\theta_n;\mathcal{D}_n) = \frac{1}{p_n} \sum_{i=1}^{p_n} \mathcal{L}(z_i^\star, y_{i,n}) \ ,
\label{eq:loss_device}
\end{equation}
where $z_i^\star$ is the solution of the fixed point equation $z_i^\star  = f_\theta(z_i^\star, x_{i,n})$, defined in relation (\ref{eq:fixed_point}) and denotes the output of the DEQ model when the input $x_n^i$ is applied.
Thus, the goal of edge device $n$ is to obtain the local DEQ model $\theta_n$ that minimizes the objective in (\ref{eq:loss_device}).

Under the FL framework, the  devices aim to collaboratively 
 train a global DEQ model, say $\theta_g$,  in a manner orchestrated by a central server. Particularly, the FL  minimizes the
aggregation of the local objectives and entails a common output for
all devices using the global model. The objective of FL is

\begin{equation}
\small
G(\theta_g) =  \sum_{n=1}^{N} w_n g_n(\theta_n;\mathcal{D}_n)
\label{eq:server_obj}
\end{equation}
\normalsize
where $w_n$ denote some weight coefficients.


\subsection{Edge device side-local update: forward and backward pass}\label{sec:3b}

The core component of the DEQ model is  the transformation $f_{\theta}(\cdot)$ in (\ref{eq:key_iter_zstar}) that is driven to equilibrium. Given this transformation, each device faces the following two challenges during its training/inference procedures. \textbf{First}, given an input $x_{i,n}$ a fixed point of $f_\theta(\cdot, x_{i,n})$ is required to be estimated efficiently during the \textit{forward pass}.  \textbf{Second}, given its dataset $\mathcal{D}_n$, each edge device needs to minimize its local objective  (\ref{eq:loss_device}) to efficiently update the weights of the local DEQ model. Without loss of generality, to simplify the notations and calculations below, we focus only on the edge device $n$ using only a single pair of training samples denoted as $\{x, y\}$.

\subsubsection{Forward pass - Calculating fixed points}\label{forward_pass}
During the training and testing procedures of edge device $n$, a large number of fixed point iterations needs to be computed based on the transformation map of DEQ model in (\ref{eq:fixed_point}) to estimate the fixed point $z^{\star} = f_\theta(z^{\star}, x)$.

A simple yet time-consuming approach to calculate the fixed point is to apply the following recursive scheme, i.e., $z^{(k+1)} = f_\theta(z^{(k)}, x)$
 until $z^{(k+1)}$ and $z^{(k)}$ are sufficiently close to each other.
In view of this, we employ  the Anderson acceleration approach \cite{walker2011anderson}, thus accelerating significantly the calculation of the fixed-point. Specifically, the Anderson acceleration method employs $m$ previous fixed point estimates to compute the next fixed point estimate, thus forming the following rule\begin{equation}
\small
    z^{(k+1)} = (1-\beta)\sum_{i=0}^{m-1}{\alpha_i z^{(k-i)}} + \beta\sum_{i=0}^{m-1}{\alpha_i f_\theta(z^{(k-i)}, x)},
\end{equation}
for $\beta>0$. The vector $\alpha \in \real^m$ is computed via 
\begin{equation}
\small
    \arg\min_{\alpha} \norm{\textbf{Q}\alpha}^2_2, \quad s.t. \quad 1^T\alpha = 1
    \label{eq:anderson}
\end{equation}
where $\textbf{Q}=[f_\theta(z^{k},x)-z^{(k)}, \dots, f_\theta(z^{k-m+1},x)-z^{(k-m+1)} ]$ is a matrix, which contains $m$ past residuals. 

\subsubsection{Backward pass - Calculating the Gradient}\label{deq_training}
The second challenge that devices encounter during their training process is the computation of backpropagation. The objective is to train their local DEQ model without the need to backpropagate through a significant number of fixed-point iterations.

Let $z^\star=f_\theta(z^\star, x)$ be the fixed point estimated during the forward pass given the input x from the local dataset of device $n$ and $l(z^\star)$ = $\mathcal{L}(z^\star, y)$ be a loss function using only an example-target $y$. The gradient with respect to the local DEQ model, denoted as $\theta$ is
\begin{equation}
\small
    \frac{\partial l}{\partial \theta} =\left(\frac{\partial z^{\star T} }{\partial \theta}\right) \left(\frac{\partial l}{\partial z^\star}\right),
\label{eq:loss_gradient}
\end{equation}
where the first factor is the Jacobian of $z^\star$ w.r.t. $\theta$ and the second factor is the gradient of the loss function. To avoid backpropagating through a large number of fixed point iterations, the implicit backpropagation \cite{NEURIPS2019_01386bd6} is performed to estimate 
the Jacobian of $z^\star$ w.r.t. $\theta$. To this end, we implicitly differentiate  both parts of  fixed point equation (\ref{eq:fixed_point}). i.e., $z^{\star} = f_\theta(z^{\star}, x)$ and we solve for $ \frac{\partial z^\star}{\partial \theta}$,  thus deriving an explicit expression for
the Jacobian
\begin{equation}
\small
    \frac{\partial z^\star}{\partial \theta} = \left(I - \frac{f_\theta(z^\star,y)}{\partial z^\star} \right)^{-1}\frac{f_\theta(z^\star,y)}{\partial \theta} 
    \label{eq:Jacobian}
\end{equation}
Using relation (\ref{eq:Jacobian}), equation (\ref{eq:loss_gradient}) is reformulated as follows
\begin{equation}
\small
    \frac{\partial l}{\partial \theta} = 
     \frac{f_\theta(z^\star,x)^T}{\partial \theta}  \left(I - \frac{f_\theta(z^\star,x)}{\partial z^\star} \right)^{-T} \frac{\partial l}{\partial z^\star}.
\label{eq:gradient2}
\end{equation}
Thus, we need to compute only a computationally efficient Jacobian-vector product, as shown in equation (\ref{eq:gradient2}) in order to obtain the gradient of the loss in (\ref{eq:loss_gradient}).
Following \cite{tutorial, my},  to calculate this Jacobian-vector product,  the  vector $\gamma$ is defined 
\begin{equation}
\small
    \gamma = \left(I - \frac{f_\theta(z^\star,y)}{\partial z^\star} \right)^{-T} \frac{\partial l}{\partial z^\star} \Rightarrow 
        \gamma = \left(\frac{f_\theta(g^\star,y)}{\partial z^\star} \right)^{T}\gamma + \frac{\partial l}{\partial z^\star}.  
            \label{eq:gamma}
\end{equation}
Note that expression (\ref{eq:gamma}) also defines a fixed point equation. Hence, solving this fixed point equation and computing the fixed point $\gamma^\star$, the gradient in 
(\ref{eq:gradient2}) can be written as
\begin{equation}
\small
    \frac{\partial l}{\partial \theta} = 
    \frac{f_\theta(z^\star,x)^T}{\partial \theta} \gamma^\star. 
    \label{eq:grad_loss}
\end{equation}
A great advantage is that  the Jacobian-vector product in (\ref{eq:gamma}) can be efficiently computed by conventional automatic differentiation tools with constant memory \cite{NEURIPS2019_01386bd6}.

\subsection{Advantages (Properties) of the DEQ models in FL}

\textbf{Communication efficiency:} One of the most important benefits that stems from the adoption of a DEQ neural network model in the FL framework, is the significant reduction in the number of model parameters that need to be transmitted.
The DEQ model utilized by edge devices can be interpreted as a compressed version of a K-layer deep network in the sense that it only employs a single layer as its central component. This layer defines the transformation function $f_\theta(\cdot)$ that drives the system towards equilibrium. 
In other words, if we consider that the transformation function $f_\theta(\cdot)$ is defined in terms of $M$ parameters, then the total number of parameters required for the transmission of a DEQ model is somewhat more than $M$, if we also consider the parameters of the output layer. On the other hand, if a $K$-layer deep neural network model was employed in the federated learning framework, then each layer typically employs a number of parameters of the order of $M$ per layer and the total number of parameters required for the transmission of one model would be close to $K\cdot M$, i.e., a multiple of what is required by adopting a DEQ model.



\textbf{Reduced memory requirements:} An obvious benefit that results from the adoption of DEQ neural network models in the federated learning framework, is that both the clients as well as the parameter server need a significantly smaller amount of memory, to store and process the models. Thus, this enables the use of devices with smaller costs. If we take into account that some models have several millions (or even billions) of parameters, we realize the importance of such a memory saving.

\textbf{Support for heterogeneous devices:} Another important advantage of the use of DEQ models in the FL framework is 
that it naturally enables the incorporation of edge devices with significantly different processing capabilities. In order to explain this property, it is important to note that a DEQ model derived by following the procedure in Section \ref{sec:3b} corresponds to an infinite depth neural network, if the algorithm used to compute the fixed point performs enough iterations to achieve convergence. Likewise, when the iterative algorithm employed by the devices to compute the fixed point of the transformation $f_\theta(\cdot)$ (see eq. \ref{eq:fixed_point}) performs a limited number of iterations, this results into a model that approximates the ideal infinite depth neural network. Thus, different devices can perform different numbers of iterations, according to their computational capabilities. Devices with more processing power can perform iterations until convergence is reached. However, less powerful clients can perform fewer iterations, at the cost of reaching approximate, but valid, models. In any case, the DEQ models derived by all heterogeneous devices can be aggregated easily at the server side, since they are all defined in terms of the same transformation $f_\theta(\cdot)$.

\subsection{Server-side - fusion rule}
On the server-side, considering that the DEQ models sent by the devices have been derived by heterogeneous edge devices that each may employ a different number of fixed point iterations, a weighted average fusion rule is proposed that effectively utilizes  this information. In more detail, at every communication round $t$, the server employs the following fusion rule, defined by 
\begin{equation}
\small
    \theta_g^{(t)} = \frac{1}{\sum_{n=1}^{N}k_n} \, \sum_{n=1}^{N} k_n \theta_n^{(t)}\ ,
\end{equation}
where $\theta_g^{(t)}$ denotes the global model, $\theta_n^{(t)}$ is the DEQ model sent to the server by client $n$ and $k_n$ denotes the number of fixed point iterations performed by client $n$.



\section{Experimental part}
In this section, we evaluate the effectiveness of the proposed Federated Deep equilibrium learning approach and compare it with the corresponding federated learning approaches using traditional deep learning models in different scenarios. The evaluation is conducted from two perspectives: 1) Communication efficiency and accuracy, and 2) Heterogeneity of edge devices in terms of their computational capabilities.

\subsection{Implementation details}
\textbf{Dataset and settings:} We utilize the well-known CIFAR10, adopting similar training/testing splits as previous studies e.g., \cite{chen2022on} and exploring two scenarios, i.e, 1) an IID data case and 2) a non IID data case. In the non IID data setting, the label ratios follow the Dirichlet distribution, in which the Dirichlet parameter was set to 0.25 \cite{chen2022on}. Since, the focus of this study is to connect the DEQ models with the federated learning, for all experiments we employed the FedAvg algorithm with a network of $N= 15$ devices.  \\
\textbf{Compared deep learning networks:} Regarding the compared methods, we consider two federated learning methods based on the FedAvg algorithm that employ two deep learning models, i.e., 1) a ResNet network and 2) a lighter CNN model. The ResNet model comprises five residual blocks and a fully connected  (FC) layer with ten neurons as outputs for ten classes. The CNN model consists of $L$ convolutional layers followed by a FC layer, where the value of the parameter $L$ varies from $4$ to $10$ throughout the experiments. \\
\textbf{The choice of DEQ model (transformation $f_\theta(.)$):} The core component of the DEQ models is the  transformation $f_\theta(.)$. In this study, we explored two DEQ models comprising transformations based on the previously mentioned ResNet and CNN networks. In the first case, since the ResNet consists of residual blocks, we considered a residual block as a transformation $f_\theta$:
\begin{equation}
\small
f_\theta(z,x) =\mathcal{B}(\sigma(z+\mathcal{B}(x+\textbf{W}_2*\mathcal{B}(\sigma(\textbf{W}_1*z)))))) \nonumber
\end{equation}
where $\mathcal{B}$ represents the Group normalization and $\sigma$ is the ReLU. Note that the DEQ model may comprise of additional residual blocks to achieve a balance between accuracy and number of parameters. In the second case, the transformation $f_\theta$ of the DEQ model comprises of a convolutional layer (or more). After finding the equilibrium point, the output of the above DEQ models is followed by a FC layer.  \\
\textbf{Parameters:} Each federated learning method utilized 90 communication rounds between the server and the edge devices. Concerning the local training process of the edge devices, we employed $5$ epochs with batch size equal to 32 using the SGD as optimizer. During the training/testing process we used $10$ fixed point iterations for the forward and backward passes.

\subsection{Results - Communication efficiency}
\begin{figure}
\centering
 \begin{subfigure}{0.23\textwidth}
     \includegraphics[width=\textwidth]{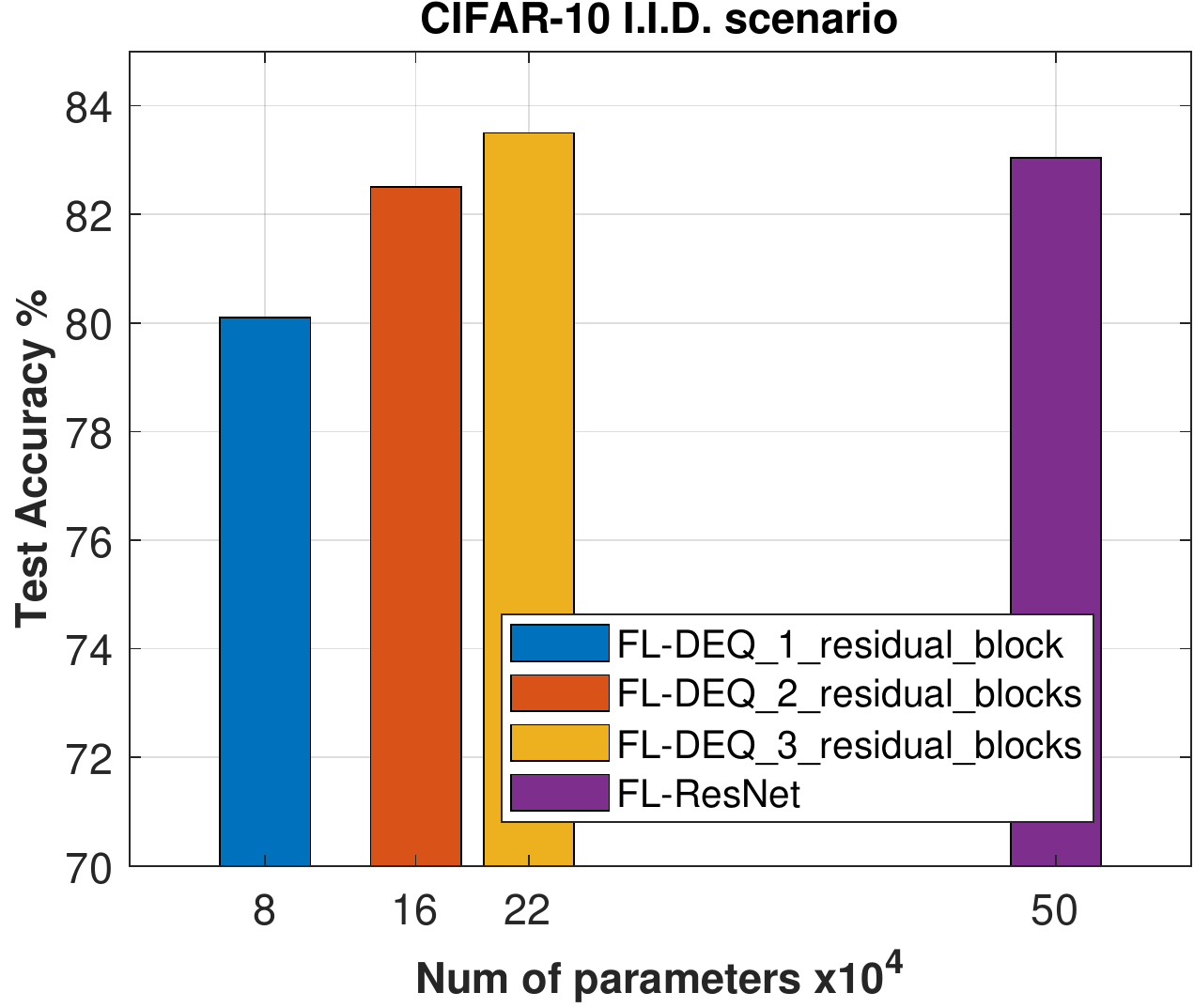}
     \caption{}
 \end{subfigure}
 \hfill
 \begin{subfigure}{0.23\textwidth}
     \includegraphics[width=\textwidth]{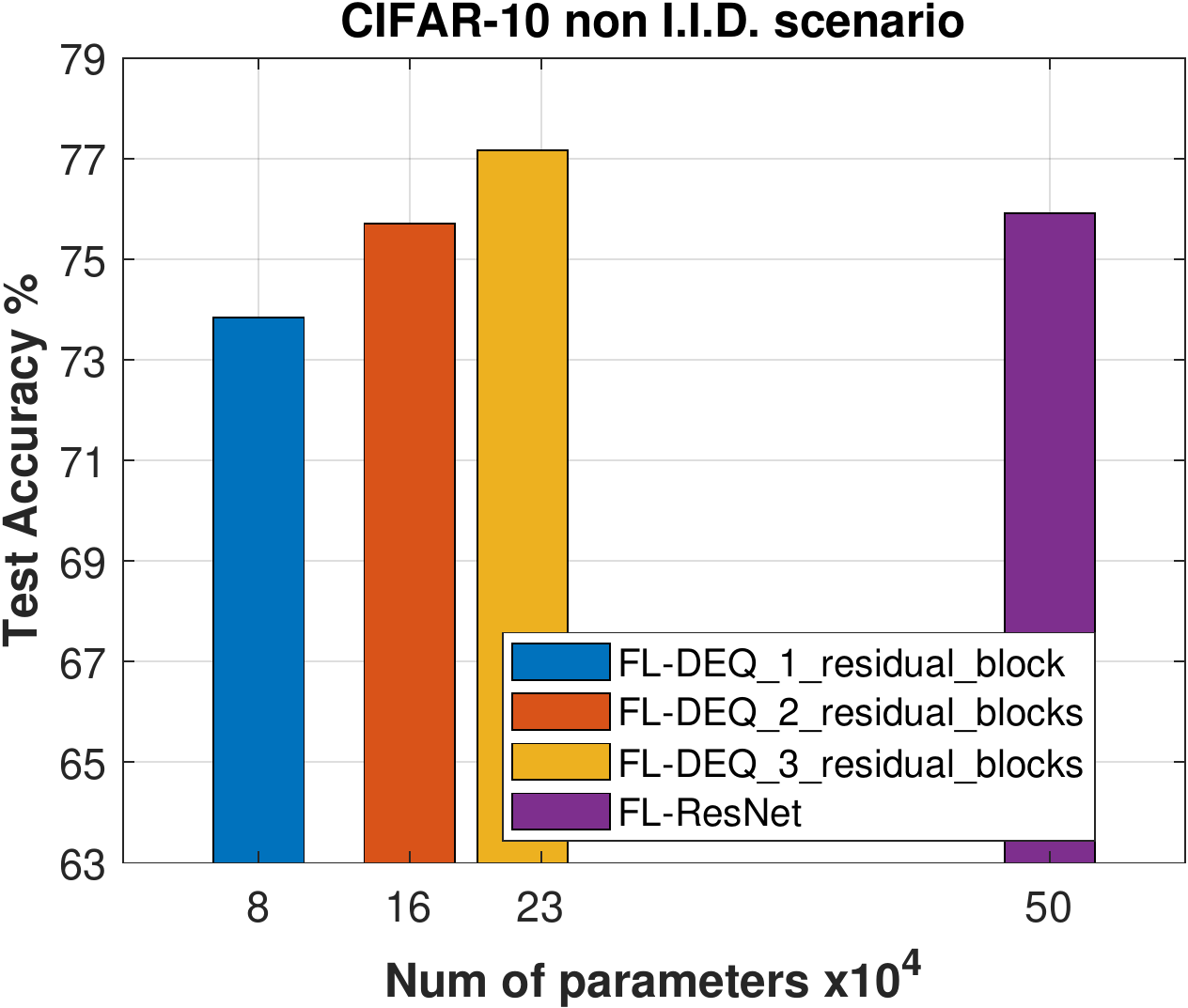}
     \caption{}
 \end{subfigure}
 \caption{The accuracy of the proposed Federated Deep Equilibrium Learning approach, which utilizes DEQ models with
several residual blocks, is compared to Federated Learning
(FL) that employs the ResNet model, for both I.I.D. and non-
I.I.D. scenarios. }
 \label{fig:resnet_accu}
\end{figure}

\begin{figure}
\centering
 \begin{subfigure}{0.23\textwidth}
     \includegraphics[width=\textwidth]{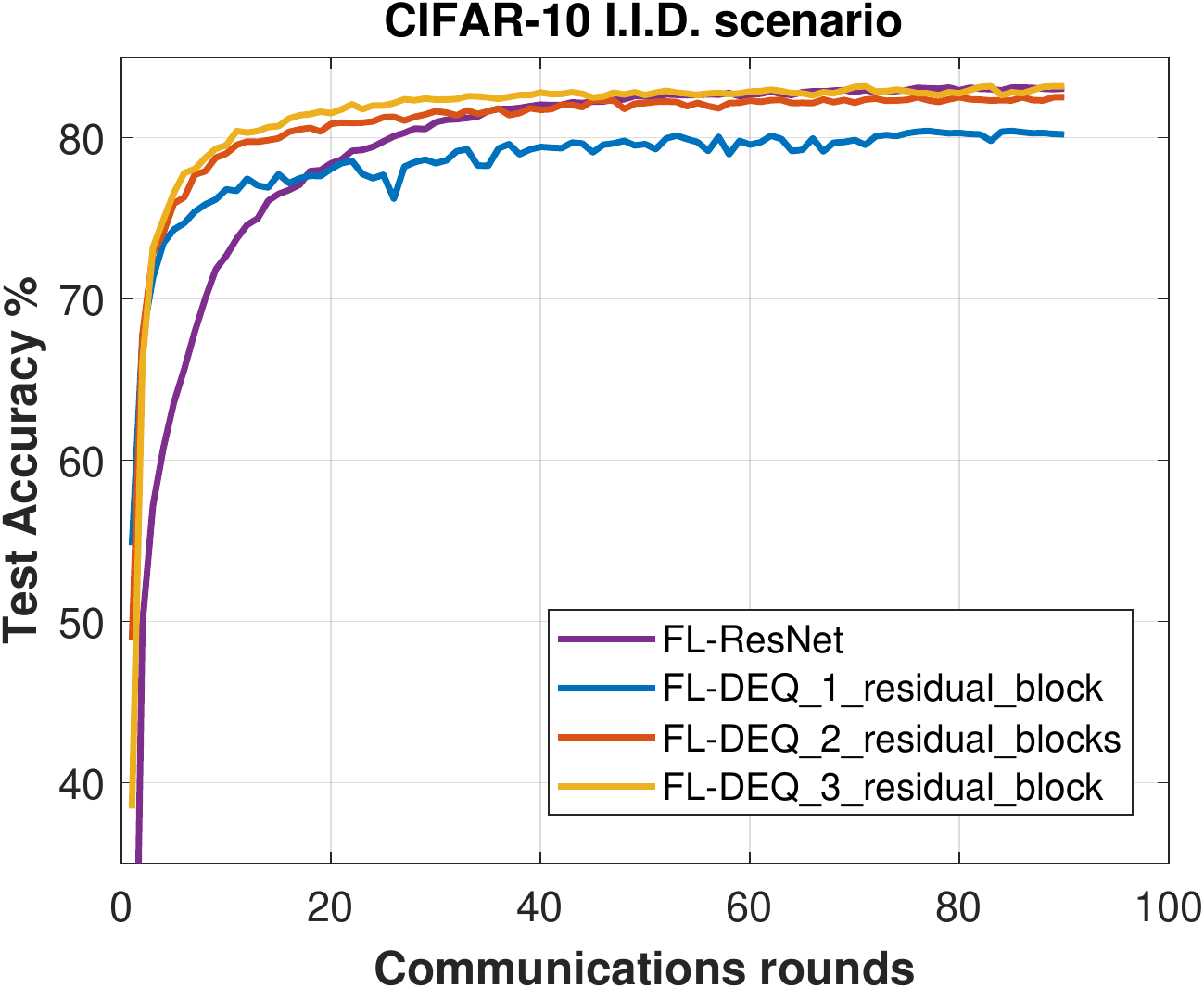}
     \caption{}
 \end{subfigure}
 \hfill
 \begin{subfigure}{0.23\textwidth}
     \includegraphics[width=\textwidth]{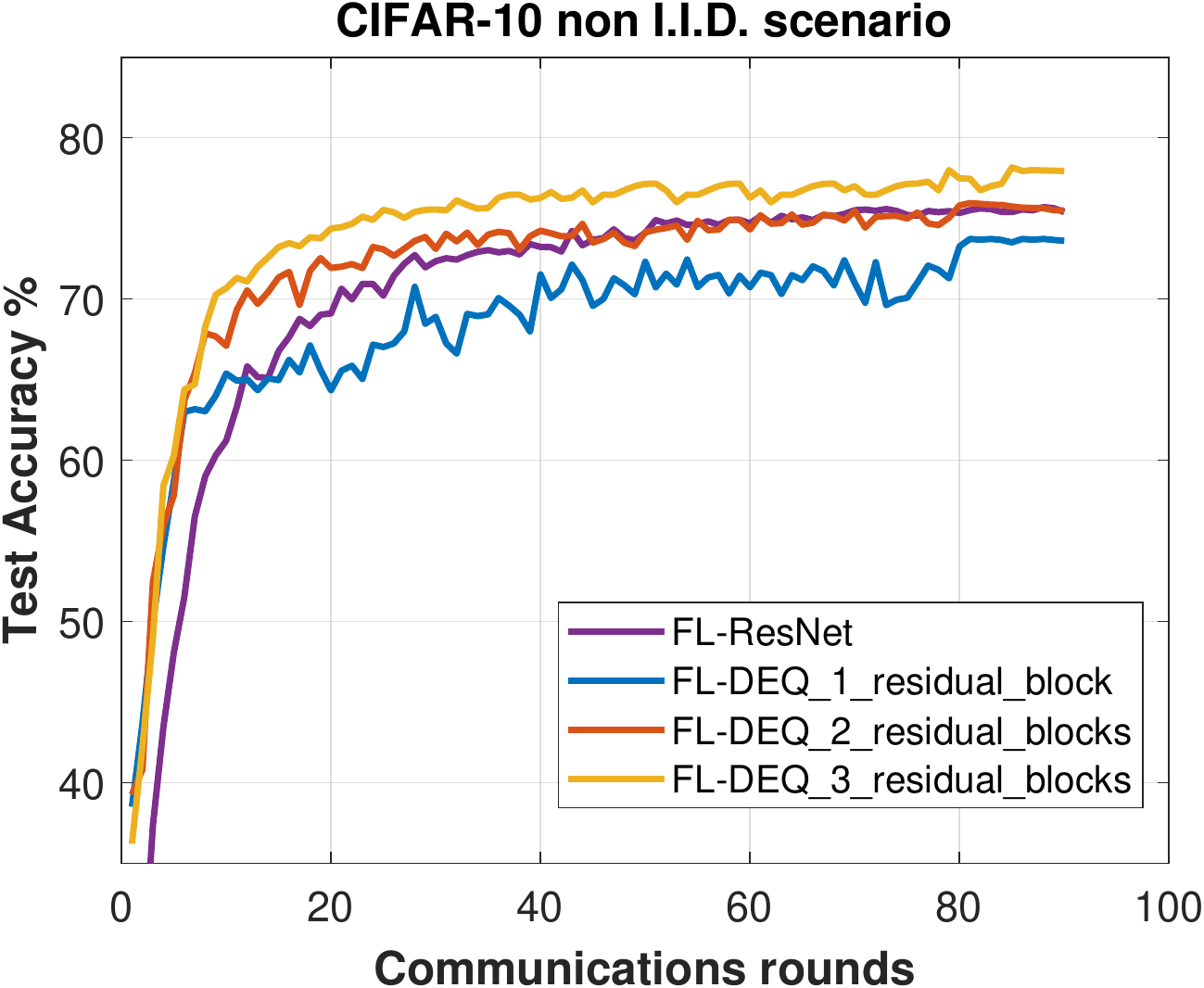}
     \caption{}
 \end{subfigure}
 \caption{The convergence of the proposed Federated Deep Equilibrium Learning approach, which utilizes DEQ models with several residual blocks, is compared to Federated Learning (FL) that employs the ResNet model, for both I.I.D. and non-I.I.D. scenarios.}
 \label{fig:resnet}
\end{figure}

\begin{figure}
\centering
 \begin{subfigure}{0.24\textwidth}
     \includegraphics[width=\textwidth]{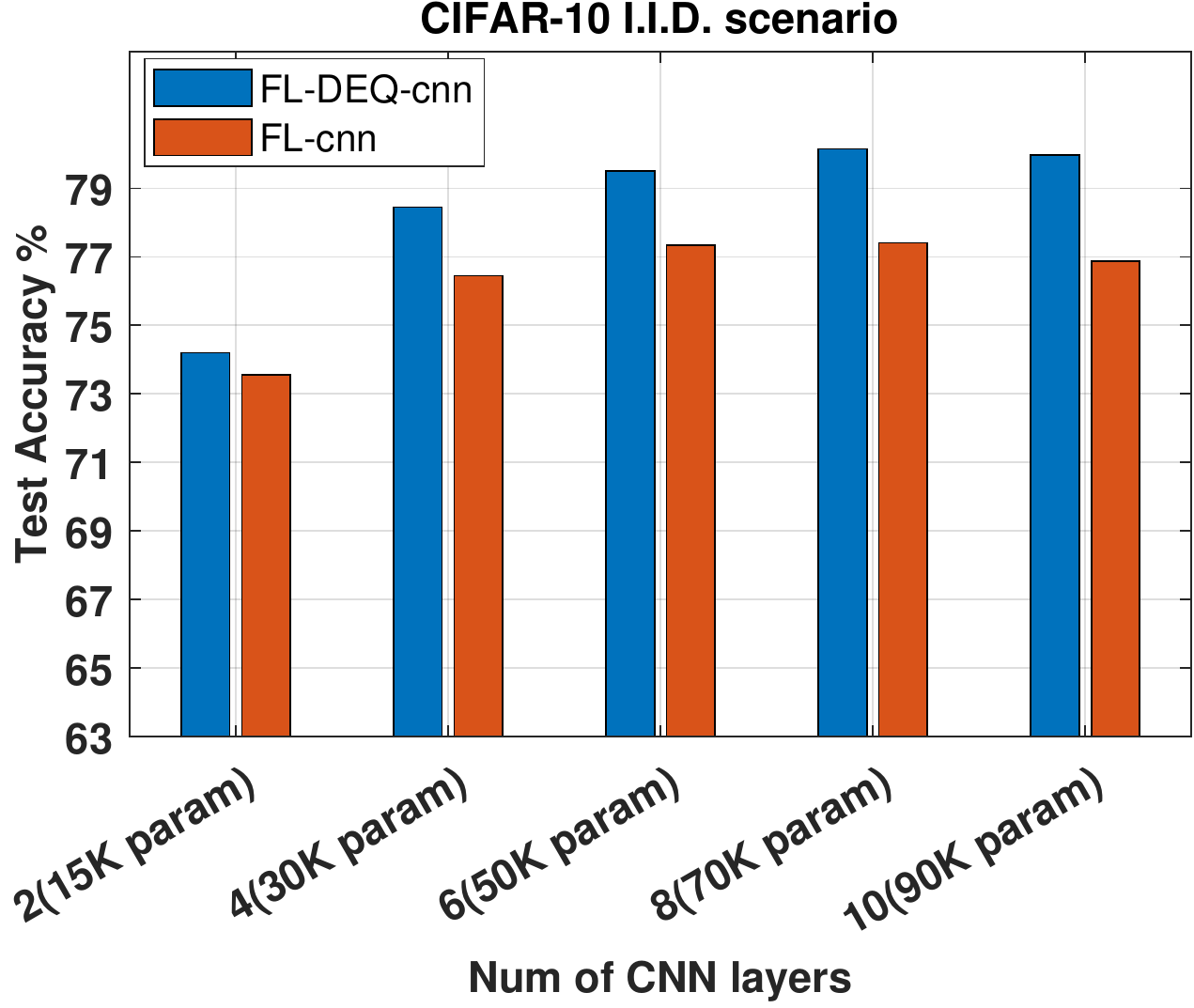}
     \caption{}
 \end{subfigure}
 \hfill
 \begin{subfigure}{0.24\textwidth}
     \includegraphics[width=\textwidth]{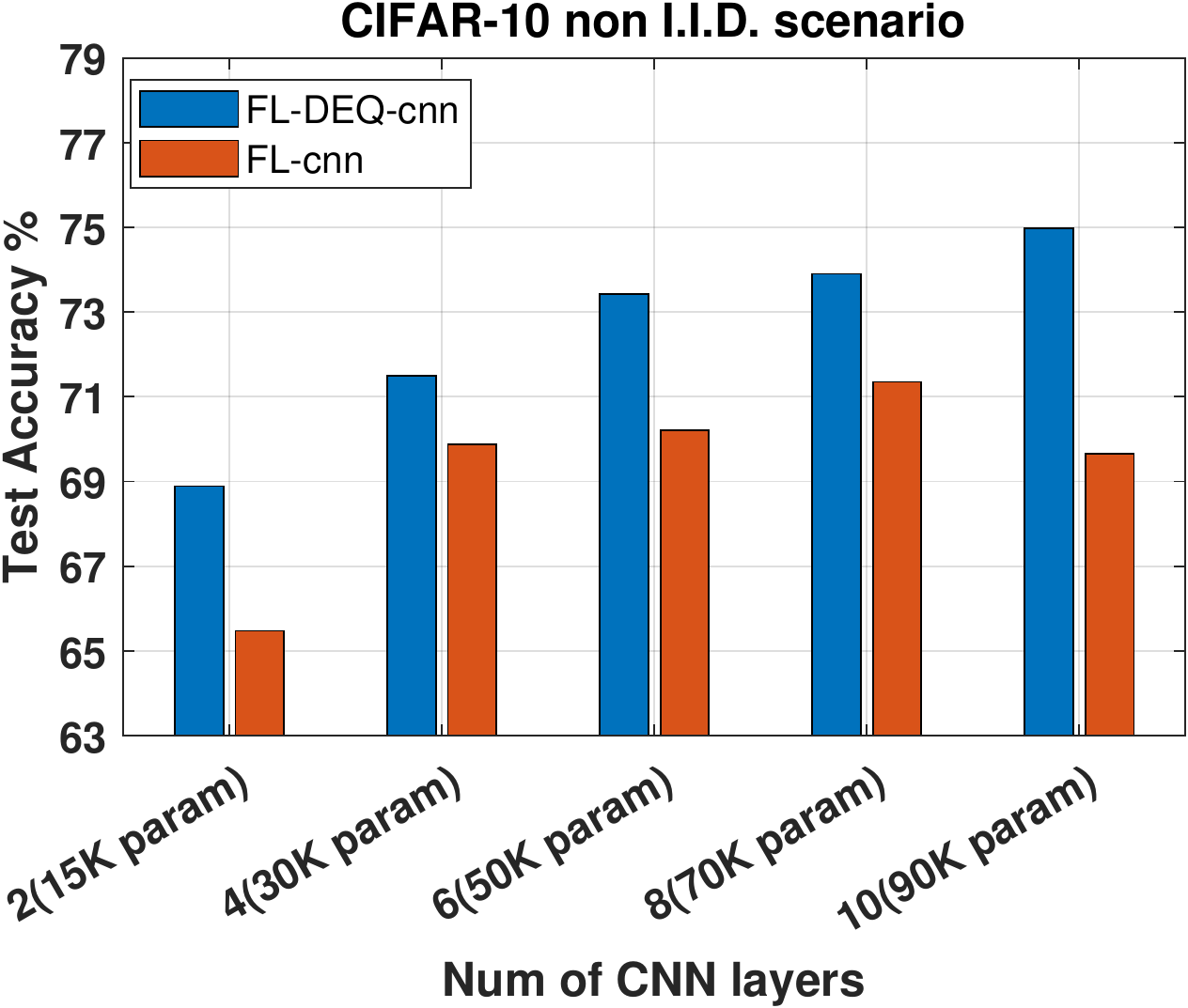}
     \caption{}
 \end{subfigure}
 \caption{The proposed Federated Deep Equilibrium Learning approach, which utilizes DEQ models with 2 to 10 convolutional layers, has been evaluated for its accuracy in comparison to Federated Learning (FL) that employs CNN models with several CNN layers, for both I.I.D. and non-I.I.D. scenarios. }
 \label{fig:cnn}
\end{figure}

Regarding the ResNet architecture, in Figure \ref{fig:resnet_accu}, we compared the FL-ResNet method with the proposed Federated Deep Equilibrium Learning approach, which utilizes a DEQ model with varying numbers of residual blocks (one, two, or three) denoted as FL-DEQ-1-residual-block, FL-DEQ-2-residual-blocks, and FL-DEQ-3-residual-blocks, respectively. The results validate that the connection of DEQ models with federated learning offers several advantages.
In more detail, utilizing only a single residual block in transformation $f_\theta$ i.e.,   the method FL-DEQ-1-residual-block, we are able to achieve more than $\mathbf{84\%}$ \textbf{reduction in the parameters} of the local models with negligible performance loss (less than $\mathbf{3.16\%}$) for both the I.I.D. and non I.I.D. scenarios as compared to the FL-ResNet approach. Interestingly, with a slight increase in the complexity of the DEQ model (using 3 residual blocks), the FL-DEQ-3-residual-blocks outperforms the FL-ResNet method, while requiring $\mathbf{56\%}$ \textbf{less parameters}. Similar results were obtained for the CNN architecture. Specifically, in Figure \ref{fig:cnn}, the Federated Deep Learning method that uses a transformation function  with $4$ convolutional layers provides a significant reduction in the number of parameters  (\textbf{more than $\mathbf{66.5\%}$}) while achieving  better performance than the all FL-CNN cases. Additionally, from Figure \ref{fig:resnet}, the three considered federated deep equilibrium learning methods demonstrate faster convergence rates requiring less communication rounds to achieve a satisfactory performance accuracy compared to  the FL-ResNet approach. 
Overall, the proposed FL approach provides significant communication gains in terms of both the number of parameters that need to be exchanged between devices and the server, and the number of communication rounds required.

\begin{figure}
\centering
 \includegraphics[scale=0.35]{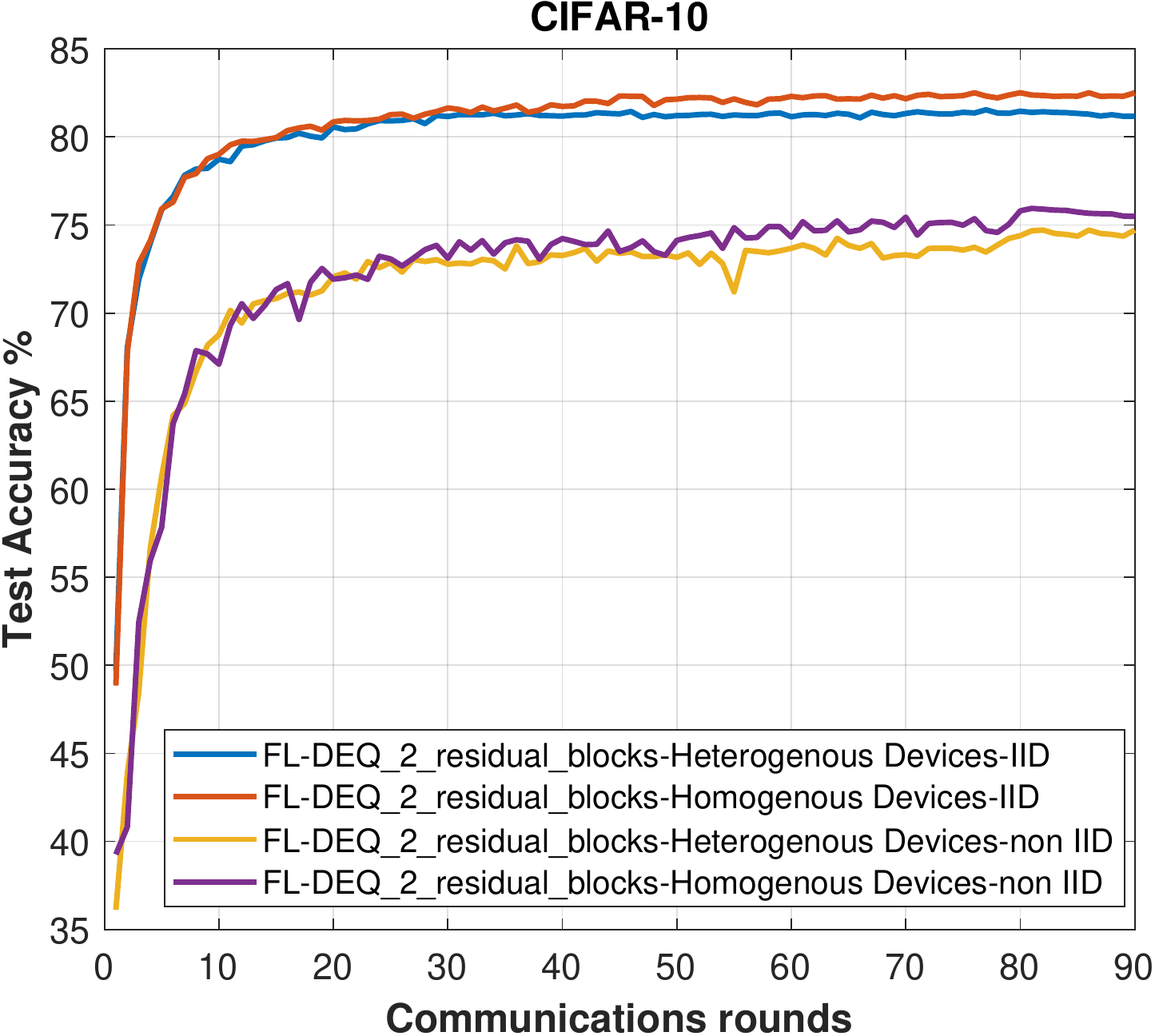}
  \caption{The FL-DEQ-2-residual-blocks method is evaluated in a heterogeneous scenario, demonstrating resilience to varying computational resources of edge devices without sacrificing convergence speed or accuracy compared to the homogeneous scenario. }
  \label{fig:heterogeneous}
\end{figure}

\subsection{Results - Heterogeneous Devices}


In this experiment, we considered a scenario in which $40 \%$ of the participating edge devices were characterized by limited computational resources, and therefore performed only $3$ fixed point iterations during their local training update. The remaining devices, with more powerful hardware, performed 10 fixed point iterations. We evaluated the performance of the FL-DEQ-2-residual-blocks method in two cases: (1) where all edge devices performed 10 fixed point iterations (homogeneous scenario), and (2) where the number of fixed point iterations varied depending on the computational resources of the edge devices (heterogeneous scenario). The results, shown in Figure (\ref{fig:heterogeneous}), demonstrate that the proposed approach is resilient to heterogeneity in the hardware capabilities of the edge devices. Specifically, the FL-DEQ-2-residual-blocks method achieved similar accuracy in both the homogeneous and heterogeneous scenarios, despite the fact that some devices performed fewer fixed point iterations than others. This suggests that the proposed approach can effectively leverage the varying computational resources of edge devices, without sacrificing accuracy or convergence speed. Note that similar results were obtained by utilizing the other proposed schemes but they have been omitted due to space limitations.

\section{Conclusions}

This study presents a novel approach that employs Deep Equilibrium  models to address several challenges in federated learning. In particular, it was shown that using DEQ models in the context of FL can effectively handle the communication overhead associated with sharing large models and the computational heterogeneity of edge devices. Promising initial experimental results were presented, indicating the potential of this approach in addressing the challenges of FL. Further experimental work is needed to fully explore the capabilities and limitations of the proposed approach, including its performance on different datasets and neural network architectures.

\bibliographystyle{IEEEtran}
\bibliography{references}

\end{document}